\documentclass[authoryear]{elsarticle}
\usepackage{lipsum}
\makeatletter
\def\ps@pprintTitle{%
 \let\@oddhead\@empty
 \let\@evenhead\@empty
 \def\@oddfoot{}%
 \let\@evenfoot\@oddfoot}
\makeatother
\usepackage[absolute,overlay]{textpos}
\usepackage{amssymb}
\usepackage{rotating}

\usepackage{paralist}
\usepackage{graphicx}
\usepackage{tabularx}
\usepackage{siunitx}
\usepackage{hyperref}
\usepackage{fancyvrb}
\usepackage{url}
\usepackage{amsmath}
\usepackage{booktabs}
\usepackage{ntheorem}
\usepackage{bm}
\usepackage{amssymb}

\usepackage{blindtext}
\usepackage{tikz}
\usepackage{lineno,hyperref}
\usepackage{subcaption}
\usepackage{booktabs}
\usepackage{framed}
\usepackage{etoolbox} 
 
\usepackage{todonotes}
\usepackage{comment}

\usepackage{paralist}
\AtBeginEnvironment{algorithmic}{\footnotesize} 

\AtBeginEnvironment{tabular}{\footnotesize} 
\usepackage{algorithm,algcompatible}
\newcommand{\mybox}[1]{%
         \begin{center}%
            \begin{tikzpicture}%
                \node[rectangle, draw=gray, top color=white!10, bottom color=white!90, rounded corners=5pt, inner xsep=5pt, inner ysep=6pt, outer ysep=10pt]{
                \begin{minipage}{0.95\linewidth}#1\end{minipage}};%
            \end{tikzpicture}%
         \end{center}%
}

\usepackage{todonotes}

\begin{document}

\begin{frontmatter}

\title{Advances in Machine Learning \\ for the Behavioral Sciences}

\author[A]{Tom{\'a}\v{s} Kliegr  \corref{cor1}}
\author[B]{\v{S}t\v{e}p{\'a}n Bahn{\'i}k}
\author[C]{Johannes F\"urnkranz}

\address[A]{Department of Information and Knowledge Engineering, Faculty of Informatics and Statistics, University of Economics Prague, Czech Republic\\
E-mail: tomas.kliegr@vse.cz}
\address[B]{Department of Management, Faculty of Business Administration,\\
University of Economics Prague, Czech Republic\\
E-mail: stepan.bahnik@vse.cz}
\address[C]{
Computational Data Analytics Group,
Department of Computer Science,\\
Johannes Kepler Universit\"at Linz,
Altenbergerstra{\ss}e 66B, 4040 Linz, Austria\\
    Email: juffi@faw.jku.at\\}
\cortext[cor1]{Corresponding author}

\begin{abstract}
The areas of machine learning and knowledge discovery in databases have considerably matured in recent years. In this article, we briefly review
recent developments as well as classical algorithms that stood the test of time. Our goal is to provide a general introduction into 
different tasks such as learning from tabular data, behavioral data, or textual data, with a particular focus on actual and potential applications in behavioral sciences. The supplemental appendix to the article also provides practical guidance for using the methods by pointing the reader to proven software implementations.  The focus is on R, but we also cover some libraries in other programming languages as well as systems with easy-to-use graphical interfaces. 

\end{abstract}


\end{frontmatter}

\begin{textblock*}{18cm}(1cm,1cm) 
   This is a post-peer-review preprint version of an article published in American Behavioral Scientist. The final version, including a freely accessible appendix with a list of software implementations, is available online at: \url{https://journals.sagepub.com/doi/full/10.1177/0002764219859639}.
   
\end{textblock*}

\section{Introduction}
Machine learning has considerably matured in recent years, and has become a key enabling technology for many data-intensive tasks. Advances in neural network-based deep learning methodologies have yielded unexpected and unprecedented performance levels in tasks as diverse as image recognition, natural language processing, and game playing. Yet, these techniques are not universally applicable, the key impediments being their hunger for data and their lack of interpretable results. These features  make them less suitable for behavioral scientists where data are typically scarce, and results that do not yield insights into the nature of the processes underlying studied phenomena are often considered of little value.

This article presents an up-to-date curated survey of machine learning methods applicable to behavioral research. Since being able to understand a model is a prerequisite for uncovering the causes and mechanisms of the underlying phenomena, we favored methods that generate \emph{interpretable models} from the multitude of those available. However, we also provide pointers to state-of-the-art methods in terms of predictive performance, such as neural networks.

Each covered method is described in nontechnical terms. To help 
researchers in identifying the best tool for their research problem, we put emphasis on examples, when most  methods covered are complemented with references to their existing or possible use in the behavioral sciences. Each described method  is supplemented with a description of software that implements it, which is provided in Supplemental Appendix (available online). Given the predominance of R as a language for statistical programming in behavioral sciences, we focus in particular on these packages.  We also cover some libraries in other programming languages, most notably in Python, as well as systems with easy-to-use graphical interfaces. 

The survey is organized by the character of input data. In the ``Tabular Data'' section, we cover structured, tabular data, for which we present an up-to-date list of methods used to generate classification models, as well as algorithms for exploratory and descriptive data mining. The ``Behavioral Data'' section covers methods and systems that can be used to collect and process behavioral data, focusing on clickstreams resulting from web usage mining, and methods developed for learning preference models from empirical data. The latter two areas can, for example, be combined for consumer choice research based on data obtained from an online retailer. Given the uptake of social media both as sources of data and objects of study, the ``Textual Data'' section provides in-depth coverage of textual data, including syntactic parsing and document classification methods used to categorize content as well as new advances that allow a representation of individual documents using word embeddings. The Internet also provides new machine-readable resources, which contain a wealth of information that can aid the analysis of arbitrary content. Knowledge graphs and various lexical resources, covered in the ``External Knowledge Sources'' section, can be used, for example, for enrichment of content of small documents (microposts), which are an increasingly common form of online communication. The ``Related Work'' section discusses related work and also covers  miscellaneous topics. such as machine learning as service systems. These can provide the behavioral scientist the ability to process very large data sets with little setup costs. The conclusion summarizes the methods covered in this article, focusing on the performance -- interpretability trade-off. It also discusses emerging trends and challenges, such as the legal and ethical dimensions of machine learning.

\section{Tabular Data}
\label{sec:tabular}

The task that has received the most attention in the machine
learning literature is the \emph{supervised learning} scenario: Given a database of
observations described with a fixed number of measurements or \emph{features} and a designated
attribute, the \emph{class}, find a mapping that is able
to compute the class value from the
feature values of new, previously unseen observations. While there
are statistical techniques that are able to solve particular
instances of this problem, machine learning techniques provide a
strong focus on the use of categorical, non-numeric attributes, and
on the immediate interpretability of the result. They also typically
provide simple means for adapting the complexity of the models to the problem at hand. This, in
particular, is one of the main reasons for the increasing popularity
of machine learning techniques in both industry and academia.

\begin{table}[tbp]
	\begin{center}
		{\small
			\begin{tabular}{ccccc}
				\emph{Education\/} & \emph{Marital Status\/} & \emph{Sex\/} & \emph{Has Children\/}
				& \emph{Approve?\/} \\ \hline
				primary    & single   & male   & no  & no \\
				primary    & single   & male   & yes & no \\
				primary    & married  & male   & no  & yes \\
				university & divorced & female & no  & yes \\
				university & married  & female & yes & yes \\
				secondary  & single   & male   & no  & no \\
				university & single   & female & no  & yes \\
				secondary  & divorced & female & no  & yes \\
				secondary  & single   & female & yes & yes \\
				secondary  & married  & male   & yes & yes \\
				primary    & married  & female & no  & yes \\
				secondary  & divorced & male   & yes & no \\
				university & divorced & female & yes & no \\
				secondary  & divorced & male   & no  & yes \\ \hline
			\end{tabular}
		}
	\end{center}
	\caption{A sample database}
	\label{tab:database}
\end{table}

Table~\ref{tab:database} shows a small, artificial sample
database, taken from \citet{jf:EJP-06}. The database contains the results
of a hypothetical survey with 14 respondents concerning the approval or
disapproval of a certain issue. Each individual is characterized by
four attributes---\emph{Education} (with possible values
\emph{primary} school, \emph{secondary} school, or
\emph{university}), \emph{Marital
	Status} (with possible values \emph{single}, \emph{married}, or
\emph{divorced}), \emph{Sex} (\emph{male} or \emph{female}), and
\emph{Has Children} (\emph{yes} or \emph{no})---that encode
rudimentary information about their sociodemographic background. The
last column, \emph{Approve?}, encodes whether the individual
approved or disapproved of the issue.

The task is to use the information in this \emph{training set}  to
derive a model that is able to predict whether a person is likely to
approve or disapprove based on the four demographic characteristics.
As most classical machine learning methods tackle a setting like this, 
we briefly recapitulate a few classical algorithms, while mentioning
some new developments as well.

\subsection{Induction of Decision Trees}

\begin{figure}[tbp]
	\centering
	\resizebox{0.4\textwidth}{!}{\includegraphics{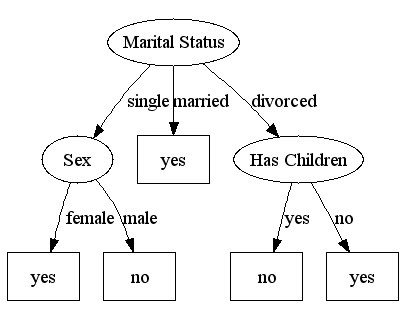}}
	\caption{A decision tree describing the dataset shown in Table~\ref		{tab:database}.}
	\label{fig:golf-dt}
\end{figure}

The induction of decision trees is one of the oldest and most popular
techniques for learning discriminatory models, which has been
developed independently in the statistical \citep{CART,CHAID} and
machine learning \citep{C4} literatures. A \emph{decision tree} is a
particular type of classification model that is fairly easy to induce
and to understand. In the statistical literature 
	\citep[cf., e.g.,][]{CART}, decision trees are also known as \emph{classification
		trees}. Related techniques for predicting numerical class values
	are known as \emph{regression trees}.
	
Figure~\ref{fig:golf-dt} shows a sample tree which might be induced
from the data of Table~\ref{tab:database}. To classify a specific instance, the decision tree  asks the question ``What is the marital status for a given instance?''. If the answer is ``married'' it assigns the class ``yes''. If the answer is divorced or single, an additional question is sought. 

In general, the classification of a new
example starts at the top \emph{node}---the \emph{root}. In our example, the root is a \emph{decision node}, which corresponds to a test of the value of the  \textit{Marital Status} attribute.
Classification then
proceeds by moving down the branch that corresponds to a particular
value of this attribute, arriving at a new decision node with a new
attribute. This process is repeated until we arrive at a terminal
node---a so-called \emph{leaf}---which is not labeled with an
attribute but with a value of the target attribute
(\textit{Approve?}). For all examples that arrive at the same leaf
value, the same target value will be predicted.
Figure~\ref{fig:golf-dt} shows leaves as rectangular boxes and decision nodes as ellipses.  

Decision trees are learned in a top-down fashion: The program
selects the best attribute for the root of the tree, splits the set
of examples into disjoint sets (one for each value of the chosen
attribute, containing all training examples that have the
corresponding value for this attribute), and adds corresponding
nodes and branches to the tree. If there are new sets that contain
only examples from the same class, a leaf node is added for each of
them and labeled with the respective class. For all other sets, a \emph{decision node} is added and associated with the best attribute for
the corresponding set as described above. 
Hence, the dataset is
successively partitioned into non-overlapping, smaller datasets
until each set only contains examples of the same class (a
\emph{pure} node). Eventually, a pure node can always be found via
successive partitions unless the training data contain two
identical but contradictory examples, that have the same
feature values but different class values.

The crucial step in decision tree induction is the choice of an
adequate attribute. 
Typical attribute selection criteria use a function that measures
the \emph{purity} of a node, that is, the degree to which the node
contains only examples of a single class. This purity measure is
computed for a node and all successor nodes that result from using
an attribute for splitting the data. The difference between the
original purity value and the sum of the values of the
successor nodes weighted by the relative sizes of these nodes, is
used to estimate the utility of this attribute, and the attribute
with the largest utility is selected for expanding the tree. The algorithm C4.5
uses information-theoretic entropy as a purity measure \citep{C4},
whereas CART uses the Gini index \citep{CART}. 
Algorithm C5.0, successor to C4.5, is noted for its best performance among all tree learning algorithms in the seminal benchmark article of \citet{fernandez2014we}.

\emph{Overfitting}  refers to the use of an overly complex model that results in worse performance on new data than would be achievable with a simpler model \cite[ch.~3]{mitchell}.  Tree models may overfit due to specialized decision nodes that refer to peculiarities of the training data.
In order
to receive simpler trees and to fight overfitting, most decision
tree algorithms  apply pruning techniques that simplify 
trees after learning by removing redundant decision nodes.

A general technique for improving the prediction quality of classifiers is to form an ensemble -- learning multiple classifiers whose individual predictions are joined into a collective final prediction. The best-known technique is \emph{random forests} \citep{RandomForests}, which uses resampling to learn a variety of trees from different samples of the data. They also use different random subsets 
of all available attributes, which not only increases the variance in the resulting trees but also makes the algorithm quite fast. However, the increased
predictive accuracy also comes with a substantial decrease in the interpretability of the learned concepts. 

\paragraph{\bf Applications in Behavioral Sciences}
Given that they are not only well-known in machine learning and data mining, but are also firmly rooted in statistics, decision trees have seen a large number of applications in behavioral sciences, of which we can list just a few.
\citet{EDM-Behavioral} provide an in-depth introduction to this family of techniques, and also demonstrate their use in a number of applications in demographic, medical, and educational areas.
In demography,
\citet{jf:EJP-06} have applied decision tree learning to the analysis of differences in the life courses in Austria and Italy, where the key issue was to model these events as binary temporal relations.
Similar techniques have also been used in survival analysis. For example, so-called survival trees
have been used in \citep{DEROSE}.
In the political sciences, decision trees have been used for 
modeling international conflicts
\citep{jf:AAI-Peace}
and international negotiation
\citep{jf:Peace-Druckman}.
\citet{rosenfeldcombining} also used decision trees to model people's behavior in negotiations.
In psychology, \citet{walsh2017predicting} used random forests to predict future suicide attempts of patients.

\subsection{Induction of Predictive Rule Sets}

Another traditional machine learning technique is the induction of rule
sets \citep{jf:Book-Nada}. The learning of rule-based models has been the main research goal
in the field of machine learning since its beginning in the early
1960s. 
Rule-based techniques have
also received some attention in the statistical community
\citep{BumpHunting}.

\begin{figure}[tbp]
	\begin{center}
		{\small
			\begin{minipage}{0.9\textwidth}
				\begin{tabbing}
					\textbf{IF} Marital Status = married \=  \textbf{THEN} yes\\
					\textbf{IF} Sex = female \> \textbf{THEN} yes\\
					\textbf{IF} Sex = male \> \textbf{THEN} no\\
					\textbf{DEFAULT} yes
				\end{tabbing}
			\end{minipage}
		}
	\end{center}
	\caption{A smaller rule set describing the dataset shown in Table~\ref
		{tab:database}}
	\label{fig:simple-golf-rules}
\end{figure}

\paragraph{Comparison between Rule and Tree Models}
Rule sets are typically simpler and more comprehensible than decision
trees, where each leaf of the tree can be interpreted as a single rule consisting of a conjunction of all conditions in the path from the root to the leaf.

The main difference between the rules generated by a decision tree
and the rules generated by a rule learning algorithm is that the
former rule set consists of non-overlapping rules that span the
entire instance space --  each possible combination of feature
values will be covered by exactly one rule. Relaxing this constraint
by  allowing potentially overlapping rules that need not span the
entire instance space may often result in smaller rule sets.

However, in this case, we need mechanisms for tie-breaking: Which rule to choose when more than one covers the given example. We also need mechanisms for default classifications: What classification to choose when no rule covers the given example. Typically, one prefers rules with a higher ratio of correctly classified examples from the training set.

\paragraph{Example of a Rule Model}
Figure~\ref{fig:simple-golf-rules} shows a particularly simple rule
set for the data in Table~\ref{tab:database}, which uses two different attributes in its first two rules. Note
that these two rules are overlapping, i.e., several examples will be
covered by more than one rule. For instance, examples 3 and 10 are
covered by both the first and the third rule. These conflicts are
typically resolved by using the more accurate rule, i.e., the rule
that covers a higher proportion of examples that support its
prediction (the first one in our case).
	Also note, that this rule set makes two
mistakes (the last two examples). These might be resolved by resorting
to a more complex rule set (such as the one corresponding to the decision tree of Figure~\ref{fig:golf-dt}),
but as stated above, it is often more advisable to
sacrifice accuracy in the training set for model simplicity to avoid
overfitting. Finally, note the default rule at the end of the rule
set. This is added for the case when certain regions of the data space
are not represented in the training set.

\paragraph{Learning Rule Models}
The key ideas for learning such rule sets are quite similar to the
ideas used in decision tree induction.  However, instead of
recursively partitioning the dataset by optimizing the purity
measure over all successor nodes (in the literature, this strategy
is also known as \emph{divide-and-conquer} learning), rule learning
algorithms only expand a single successor node at a time, thereby
learning a complete rule that covers part of the training data.
After a complete rule has been learned, all examples that are
covered by this rule are removed from the training set, and the
procedure is repeated with the remaining examples. This strategy is
also known as \emph{separate-and-conquer} learning. 
Again, pruning
is a good idea for rule learning, which means that the rules only
need to cover examples that are \emph{mostly} from the same class.
It turns out to be advantageous to prune rules immediately after
they have been learned  before successive rules are learned
\citep{jf:MLJ}.

The idea to try to prune or simplify each rule right after it has been learned has been exploited in the well-known RIPPER algorithm  \citep{Cohen:1995:FER:3091622.3091637}. This algorithm has been frequently used in applications because it learns very simple and understandable rules.
It also added a postprocessing phase for optimizing a rule set in the context of other rules. The key idea is to remove one rule out of a previously learned rule set and try to relearn the rule in the context of previous rules and subsequent rules. Another type of approach to rule learning, which heavily relies on effective pruning methods, is Classification Based on Associations \citep{Liu98integratingclassification} and succeeding algorithms. Their key idea is to use algorithms for discovering association rules (cf.~``Discovering Interesting Rules'' section), and then combine a selection of the found rules into a predictive rule model.\footnote{The Classification Based on Associations algorithm does not generate a rule set but a rule list. The difference is that in a predictive rule list, the order of rules is important as it signifies precedence.}

\paragraph{Current Trends}
Current work in inductive rule learning is focused on finding simple rules via optimization \citep{Rules-ColumnGeneration,RuleSets-BayesianOptimization,Rules-MaxSAT}, mostly with the goal that simple rules are more interpretable. However, there is also some evidence that shorter rules are not always more convincing than more complex rules \citep{jf:CognitiveBias-Interpretability,jf:DS-16-ShortRules}.
Another line of research focuses on improving the accuracy of rule models, often by increasing their expressiveness through fuzzification, i.e., by making the decision boundary between different classes softer. 
 At the expense of lower interpretability, fuzzy rule learning algorithms such as SLAVE \citep{garcia2014overview}, FURIA \citep{huhn2009furia} and FARC-HD \citep{alcala2011fuzzy} often outperform models with regular, ``crisp'' rules.

\paragraph{\bf Applications in Behavioral Sciences}
Similar to decision trees, rule learning can be generally used for prediction or classification in cases where interpretability of the model is important. Rule learning could also be useful in domains where the output of the model should be easily applicable for a practitioner, such as a physician or a psychologist, given that the resulting model can be easier to remember and apply than a logistic regression or a decision-tree model.

Multiple studies used the  RIPPER  algorithm \citep{Cohen:1995:FER:3091622.3091637}, which
is considered to be the state-of-the-art in
inductive rule learning, for learning classification rules. Classification rules may be used for classification of documents in various categories. For example, one study \citep{STUMPF2009639} used RIPPER and other algorithms to classify emails. The RIPPER algorithm outperformed Naive Bayes, another popular machine learning algorithm, in terms of classification accuracy. Furthermore, rule-based explanations were considered, on average, the most understandable, which might be especially useful when the interpretation of the output of the algorithm or further work with the algorithm's results is necessary. 


%
Other uses of RIPPER include 
 classifying the strengths of opinions in  nested clauses \citep{wilson2004just} and predicting students' performance \citep{kotsiantis2002efficiency}. Some of the studies using decision trees are also used for rule learning \citep{jf:AAI-Peace,jf:EJP-06}. 

Rule learning is suggested as a possible computational model in developmental psychology \citep{shultz2013computational}. These algorithms, or decision tree models convertible to rules, could,  therefore, be used in psychology to simulate human reasoning.


\subsection{Discovering Interesting Rules}
\label{sec:interesting-rules}
The previous section focused on the use of rules for prediction, but rule learning can be also adapted for exploratory analysis, where only rules corresponding to interesting patterns in data are generated. 

A commonly used approach for this task is association rule learning. Algorithms belonging to this family are characterized by outputting all rules that match user-defined constraints on interestingness. These constraints are called interest measures and are typically defined by two parameters: minimum confidence threshold and minimum support threshold.

If we consider rule  {\small r: \textbf{IF} Antecedent \textbf{THEN} Consequent }, then \emph{rule confidence} is the proportion of objects correctly classified by the rule to all objects matched by the antecedent of the rule.  An object is correctly classified when it matches the entire rule (its antecedent and consequent), and incorrectly classified if it matches only the antecedent, but not consequent.   \emph{Rule support}  is typically defined as the proportion of objects correctly classified by the rule to all objects in the training data.

\mybox{\noindent\textbf{Example. } 
Let us consider the following object

\centerline{ $o=\{income=low, district=London,  savings=high, risk=low\}$ }

and rule \emph{r}:
{\small\textbf{IF} income=low \textbf{AND} district=London} \textbf{THEN} \emph{risk=high}.

Object $o$ matches rule $r$, because $o$ meets all conditions in the antecedent of  $r$. Rule $r$ will incorrectly classify $o$, because the class assigned by rule consequent does not match the value of the target attribute \emph{risk} of  $o$. }

Apriori \citep{agrawal1993mining} is the most well-known algorithm for mining association rules. There are also newer algorithms, such as FP-Growth, which can provide faster performance.
While association rule mining is commonly used for discovering interesting patterns in data, the simplicity of the generated rules as well as restricted options for constraining the search space may become a limitation. 

One common problem with the application of association rule mining stems from the fact that all rules matching user-defined interestingness thresholds are returned. There may be millions of such rules even for small datasets, resulting in impeded interpretability of the resulting list of rules. A possible solution is to apply \emph{pruning}, which will remove redundant rules. Another limitation of association rule mining is a lack of direct support for numeric attributes.

An alternative approach to pruning is to better focus the generation of association rules. This approach is provided by the  GUHA method \citep{hajek2010guha}, which was initially developed with the intent to automatically search for all statistical hypotheses supported by data.   The method enables the user with many  fine-grained settings for expressing what should be considered as an interesting hypothesis. 
The trade-off is that GUHA has a slower performance on larger datasets compared with association rule mining performed with Apriori  \citep{rauch2017apriori}.

Another related task applicable to descriptive and explorative data mining is \emph{subgroup discovery}, which finds groups of instances in data, which exhibit ``distributional unusualness with respect to a certain property of interest'' \citep{wrobel1997algorithm}. 
A number of \emph{quality measures} were developed for subgroup discovery, but  interest measures applied in association rule mining can be used as well.  By choosing a suitable quality measure, the subgroup discovery task can thus be adapted for a range of diverse goals, such as  mining for unexpected patterns. A subgroup can be considered as unexpected when it significantly deviates from the total population in terms of the selected quality measure \citep{atzmueller2015subgroup}.

Subgroup discovery approaches are algorithmically  diverse, with both  association rule mining and predictive rule learning algorithms used as a base approach \citep{herrera2011overview}. 
The use of  subgroup discovery can be considered over association rule mining when the task at hand involves a numeric target attribute. Some subgroup discovery algorithms also address the problem of too many rules generated by the convenient \emph{top-k} approach, which returns only $k$ top subgroups according to the selected quality metric.

\paragraph{\bf Applications in Behavioral Sciences}
Association rule mining has been extensively used to find interesting patterns in data in a number of disciplines.  Selected recent applications include exploration of mathematics anxiety among engineering students \citep{10.1007/978-3-642-22191-0_43} or discovering color-emotion  relationships \citep{feng2010using}. An interdisciplinary review of applications of subgroup discovery is provided by \citet{herrera2011overview}. More recently,  subgroup discovery was used, for example, to study the relationship between technology acceptance and personality by \citet{10.1007/978-3-642-34347-6_16}. \citet{Goh2007} provide an accessible introduction to association rule mining  aimed at behavioral researchers. 




\subsection{Neural Networks and Deep Learning}

Neural networks have a long history in artificial intelligence and machine learning. First works were motivated by the attempt to model neurophysiological insights, which resulted in mathematical models of neurons,
so-called perceptrons \citep{Perceptron-Book}. Soon, their limitations were recognized \citep{Perceptron-Minsky}, and interest in them subsided until \citet{Backpropagation} introduced backpropagation, which allows to train multi-layer networks effectively. While a perceptron can essentially only model a linear function connecting various input signals $x_i$ to an output signal
\begin{math}
  f(\mathbf x) = \sum_i w_i \cdot x_i
\end{math}
by weighting them with weights $w_i$,
multi-layer networks put the linear output through non-linear so-called \emph{activation functions}, which allow one to model arbitrary functions via complex neural networks \citep{NN-Approximation}. This insight led to a large body of research in the 1990s, resulting in a wide variety of applications in industry, business, and science \citep{NN-Applications} before the attention in machine learning moved to alternative methods such as support vector machines.

Recently, however, neural networks have surfaced again in the form of so-called deep learning, which often leads to better performance \citep{DeepLearning,DeepLearning-Nature,DeepLearning-Overview}. 
Interestingly, the success of these methods is not so much based on new insights---the key methods have essentially been proposed in the 1990s---but on the availability of huge labeled datasets and powerful computer hardware that allows their use for training large networks. 

The basic network structure consists of multiple \emph{layers} of fully connected nodes. Each node in layer $L_{i+1}$ takes the outputs of all nodes in layer $L_i$ as input. For training such networks, the input signals are fed into the input layer $L_0$, and the output signal at the last layer $L_l$ is compared to the desired output. The difference between the output signal and the desired output is propagated backward through the network, and each node adapts the weights that it puts on its input signals so that the error is reduced. For this adaptation, error gradients are estimated, which indicate the direction into which the weights have to be changed in order to minimize the error. These estimates are typically not computed from single examples, but from small subsets of the available data, so-called mini-batches. Several variants of this stochastic gradient descent algorithm have been proposed with AdaGrad \citep{AdaGrad} being one of the most popular ones. Overfitting the data has to be avoided with techniques such as dropout learning, which in each optimization step randomly exempts a fraction of the network nodes from training \citep{Dropout}.

Multiple network layers allow the network to develop data abstractions, which is the main feature that distinguishes deep learning from alternative learning algorithms. This is  most apparent when auto-encoders are trained, where a network is trained to map the input data upon itself but  is forced to project them into a lower-dimensional \emph{embedding space} on the way \citep{Denoising-Autoencoder}.

In addition to the conventional fully connected layers, there are various special types of network connections. For example, in computer vision, \emph{convolutional layers} are commonly used, which train multiple sliding windows that move over the image data and process just a part of the image at a time, thereby learning to recognize local features. These layers are subsequently abstracted into more and more complex visual patterns \citep{ConvNN-ImageNet}. For temporal data, one can use \emph{recurrent neural networks}, which do not make predictions for individual input vectors, but for a sequence of input vectors. To do so, they allow feeding abstracted information from previous data points forward to the next layers. A particularly successful architecture are LSTM networks, which allow the learner to control the amount of information flow between successive data points \citep{LSTM}.

The main drawback of these powerful learning machines is the lack of interpretability of their results. Understanding the meaning of the generated  variables is
crucial for transparent and justifiable decisions.  
 Consequently, the interest in methods that make learned models more interpretable has increased with the success of deep learning.
Some research has been devoted to trying to convert
such arcane models to more interpretable rule-based 
\citep{NN-RuleExtraction}
or tree-based models	\citep{NN-SoftDecisionTrees},
which may be facilitated with appropriate neural network training techniques \citep{jf:DS-17-DNN-Retraining}.
Instead of making the entire model interpretable, methods like LIME
\citep{LIME} are able to provide local explanations for
inscrutable models, allowing a trade-off between fidelity to the original
model with interpretability and complexity of the local model.
There is also research on developing alternative deep learning methods, most notably sum-product networks \citep{SPN-Interpretation}. These methods are firmly rooted in probability theory and graphical models and are therefore easier to
interpret than neural networks.

\paragraph{\bf Applications in Behavioral Sciences} 
Neural networks are  studied and applied in  psychological research within the scope of  \emph{connectionist models} of human cognition since about 1980s \citep{houghton2004introduction}.
The  study of artificial neural networks in this context  has intensified in recent years in response to algorithmic advances. \citet[p.~467]{mckay2017treatments} review approaches involving artificial neural networks for studying psychological problems and disorders. For example,  schizophrenic thinking is studied by  purposefully damaging artificial neural networks.
Neural networks have also been used to study non-pathological aspects of human decision making, such as consumer behavior \citep{greene2017neural}.


Deep neural networks have enjoyed considerable success in areas such as computer vision \citep{ConvNN-ImageNet}, natural language understanding \citep{DL-NLP}, and game-playing \citep{AlphaGo}. However, these success stories are based on the availability of large amounts of training data, which may be an obstacle to  wide use in behavioral sciences.


\section{Behavioral Data}
\label{sec:behavioral}
Machine learning and data mining have developed a variety of methods for analyzing behavioral data, ranging from mimicking behavioral traces of human experts, and are also known as behavioral cloning \citep{BehavioralCloning}, to the analysis of consumer behavior in the form of recommender systems \citep{RecSys-Introduction}. 
In this section, we will look at two key enabling technologies,  the analysis of log data and the analysis of preferential data.

\subsection{Web Log and Mobile Usage Mining}
Logs of user interactions with web pages and mobile applications can serve as a trove of data for psychological research seeking to understand, for example, consumer behavior and information foraging strategies.
The scientific discipline providing the tools and means for studying this user data in the form of \emph{click streams} is called \emph{web usage mining} \citep{liu2011web}. Many web usage mining approaches focus on the acquisition and preprocessing of data. These two steps  are also the main focus of this section.

\begin{figure}
  \includegraphics[width=\linewidth]{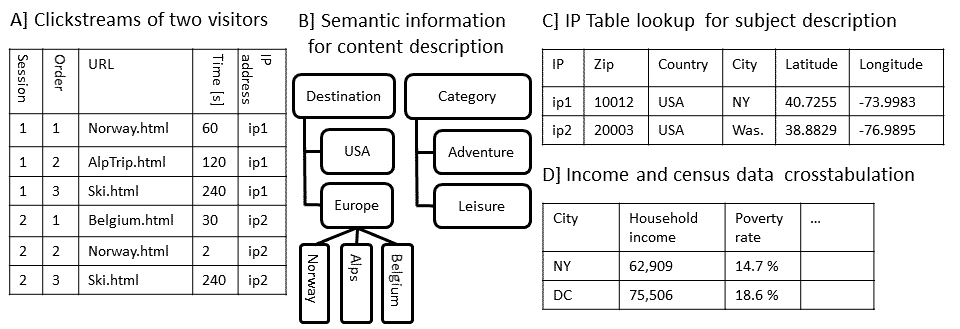}
  \caption{Data collection for web usage mining}
  \label{fig:wum}
\end{figure}

\emph{Data Collection.}
For web usage mining, there are principally two ways of collecting user interactions. Historically, the administrators of servers where the web site is hosted were configuring the server in such a way that each request for a web page was logged and stored in a text file. Each record in this \emph{web log} contains information such as the name of the page requested, timestamp, the IP address of the visitor, name of the browser, and  resolution of the screen,  providing input for web usage mining.
An alternative way is to use \emph{Javascript trackers}  embedded in all web pages of the monitored web site instead of web logs. When a user requests the web page, the script is executed in the user's browser. It can collect similar types of information as web logs, but  the script can also interact with the content of the page, acquiring the price and category of the product displayed. The script can be extended to track user behavior within the web page, including mouse movements. This information is then typically sent to a remote server, providing \emph{web analytics as a service}.
In general, Javascript trackers provide mostly  advantages over web logs as they can collect more information and are easier to set up and operate. 
Figure~\ref{fig:wum}A presents  an example of a clickstream collected from a travel agency website and Figure~\ref{fig:wum}B shows the additional information about the content of the page, which can be sent by the Javascript tracker.

\emph{Data Enrichment.}
In addition to user interactions, data collection may involve obtaining a semantic description of data being interacted with, like price and category of a product.  This information can be sent by the tracked web page. When this is not possible, one can resort to using web crawlers and scrapers. \emph{Web crawler} is software which downloads web pages and other content from a given list of web sites and stores them in a database. \emph{Web scrapers} provide means of subsequent processing of the content of web pages. This software  provides a description of information to look for, such as prices  or product categories, finds the information on the provided web page, and saves it in a structured way to a database.

Further enrichment of data can be performed, for example, through mapping IP addresses to regions via dedicated databases and software services. Their outputs include, among other information, zip codes, which might need to be further resolved to variables meaningful for psychological studies. This can be achieved using various openly accessible datasets. For example, for the United States, there is the Income tax statistics dataset\footnote{\url{https://catalog.data.gov/dataset/zip-code-data}}, which maps zip codes to several dozen income-related attributes. Other sources of data include \url{https://datausa.io/} and \url{https://data.world}. This enrichment is exemplified in Figure~\ref{fig:wum}C-D.

\emph{Data Preprocessing and Mining.}
The output of the data collection phase for web usage mining can be loosely viewed as a set of $n$ \emph{user interactions}. User interactions that take place within a given time frame (such as 30 minutes) are organized into \emph{sessions}. Each user interaction is also associated with a unique user identifier. When \emph{web logs} are used, individual records may need to be grouped into sessions by a heuristic algorithm, possibly resulting in some errors. On the other hand, records are naturally grouped into sessions when Javascript-based trackers are used.

Clickstream data are in a sequential format, in which, for example, \emph{sequential patterns or rules} \citep{agrawal1995mining} can be discovered.  

\mybox{\noindent\textbf{Example. } 

Considering the input presented in Fig.~\ref{fig:wum}A and a minimum support threshold of 30\%,  the maximum gap between two sequences = 2 and a minimum confidence of 50\%, the list of  discovered sequential rules includes:

\vspace{3mm}
{\small
\textbf{IF} Norway.html,  AlpTrip.html \textbf{THEN} Ski.html, conf = 100\%, supp =50\%.}
\vspace{3mm}

This rule says that in all (100\%) sessions where the user  visited  Norway.html and later AlpTrip.html, the user later also visited Ski.html. The number of sessions complying to this rule amounted to 50\% of all sessions.
}

Note that  the elements in the consequent of a sequential rule occur at a later time than the elements of the antecedent. 
As shown in \citep[p.~540-543]{liu2011web}, the sequential representation can also be transformed to a tabular format, which allows for the application of many standard implementations of machine learning algorithms.

\paragraph{\bf Applications in Behavioral Sciences} The use of clickstreams has a direct application in the study of consumer behavior. For example, \citet{senecal2005consumers} examined the use of product recommendations in online shopping. Other related research involves using various cognitive phenomena to explain the effect of online advertisements \citep{rodgers2000interactive}, determine the visitor's intent \citep{moe2003buying}, or  analyze reasons for impulse buying on the Internet \citep{koski2004impulse}. However, the use of data from web sites does not have to be limited to the study of consumer behavior. 
For example,  primacy and recency effects were used to explain the effect of link position on the probability of a user clicking on the link \citep{murphy2006primacy}. 
Process tracing methods have a rich history in the study of decision making and some methods, for example, mouse tracking analysis \citep{stillman2018mouse}, can be easily employed with data from Javascript trackers.




\subsection{Preference Learning}

\emph{Preference learning} is a recent addition to the suite of
learning tasks in machine learning \citep{plbook}. Roughly speaking,
preference learning is about inducing predictive preference models
from empirical data, thereby establishing a link between machine
learning and research fields related to preference modeling and
decision making. The key difference to conventional supervised
machine learning settings is that the training information is
typically not given in the form of single target values, like in
classification and regression, but instead in the form of pairwise
comparisons expressing \emph{preferences} between different objects or
labels. 
%

In general, the task of preference learning is to rank a set of objects based on observed preferences. The ranking may also depend on a given context. For example, the preference between red wine or white wine for dinner often depends on the meal one has ordered. Maybe the best-known instantiation of preference learning are \emph{recommender systems} \citep{RecSys-Introduction,plbook:Gemmis}, which solve the task of ranking a set of products based on their interest for a given user. In many cases, neither the products nor the user is characterized with features, in which case the ranking is based on similarities between the recommendations across users (user-to-user correlation) or items (item-to-item correlations) \citep{PredictiveCollaborativeFiltering}. 
In many cases, we can observe features of the context, but the objects are only designated with unique labels. This task is also known as \emph{label ranking}; \citep{plbook:Vembu}.
In \emph{object ranking}, on the other hand, the objects are described with features, but there is no context information available \citep{plbook:Kamishima-1}. Finally, if both the contexts and the objects are characterized with features, we have the most general ranking problem, \emph{dyad ranking} \citep{DyadRanking-PL}, where a set of objects is ranked over a set of different contexts. The best-known example is the problem of learning to rank in Web search where the objects are web pages,  the contexts are search queries, and the task is to learn to rank Web pages according to their relevance to a query.

Preferences are typically given in the form of pairwise comparisons between objects.
Alternatively, the training information
may also be given in the form of (ordinal) \emph{preference degrees}
attached to the objects, indicating an \emph{absolute} (as opposed to a relative/comparative)
assessment. 

There are two main approaches to learning representations of
preferences, namely \emph{utility functions}, which
evaluate individual alternatives, and \emph{preference relations},
which compare pairs of competing alternatives.
From a machine learning
point of view, the two approaches give rise to two different kinds of learning.
The latter, learning a preference relation, deviates more strongly
from conventional problems like classification and
regression, as it involves  prediction of complex structures, such
as rankings or partial order relations, rather than a prediction of single
values. Moreover, training input in preference learning will not be offered in the form
of complete examples, as is usually the case in supervised learning, but it may comprise more general types of
information, such as relative preferences or different kinds of
indirect feedback and implicit preference information.
On the other hand, the learning of a utility function, where the preference information is used to learn a function that assigns a numerical score to a given object, is often easier to apply because it enforces transitivity on the predicted rankings.

\paragraph{\bf Applications in Behavioral Sciences} 
For many problems in the behavioral sciences, people are required to make judgments about the quality of certain courses of action or solutions. However, humans are often not able to determine the precise utility value of an option, but they are typically able to compare the quality of two options.  Thurstone's \emph{Law of Comparative Judgment}
essentially states that such pairwise comparisons correspond to an
internal, unknown utility scale \citep{Thurstone}. Recovering this
hidden information from such qualitative preference is studied in
various areas such as ranking theory \citep{AnalyzingRankData}, social choice
theory \citep{Preferences-ShortIntro}, voting theory
\citep{ProbabilisticVotingTheory}, 
sports \citep{WhosNo1}, negotiation theory \citep{druckman1993situational}, decision theory
\citep{MCDA}, democratic peace theory \citep{cuhadar2014representative}, and marketing research \citep{ConjointAnalysis}. 
%
Thus, many results in preference learning are based on established statistical models for ranking data, such as the 
Plackett-Luce \citep{Plackett,Luce} or Bradley-Terry \citep{BradleyTerry} models, which allow an analyst to model probability distributions over rankings.

Given that preference and ranking problems are ubiquitous, computational models for solving such problems can improve prediction and lead to new insights. For example, in voting theory and social choice,
\citet{ParliamentaryVoting-Problems} use computational methods to analyze several parliamentary voting procedures.








\section{Textual Data}
\label{sec:text}
Much data analyzed in the behavioral sciences take the form of text. The rise of online communication has dramatically increased the volume of textual data available to behavioral scientists. In this section, we will review methods developed in computational linguistics and machine learning that can help the researcher to sift through textual data in an automated way. These methods increase the scale at which data can be processed and improve the reproducibility of analyses since a subjective evaluation of a piece of text can be replaced by automated processing, which produces the same results given the same inputs.

We review various methods for representing text with vectors, providing a gateway for further processing with machine learning algorithms. This is followed by methods for text annotation, including additional information, such as parts of speech for individual words or the political orientation of people mentioned in the text. The section concludes with machine learning algorithms for document classification, which operates on top of the vector-based representation of text.

\subsection{Word Vectors and Word Embeddings}
\label{ss:wordvectors}
A \emph{vector space model} was developed to represent a document in the given collection as a point in a space \citep{turney2010frequency}. The position of the document is specified by a vector, which is typically derived from the  frequency of occurrence of individual words in the collection. The notion of vector space models was further extended to other uses, including representation of words using their context.

Vector-based representation  has important psychological foundations \citep{Hinton:1986:DR:104279.10428,turney2010frequency}. Word vectors closely relate to a \emph{distributed representation};  
that is, using multiple (reusable) features to represent a word. \citet{landauer2013handbook} provide further  empirical and theoretical justification for psychological plausibility of selected vector space models.

 
There are multiple algorithms that can be applied to finding word vectors. They have a common input of an unlabeled collection of documents, and their output can be used to represent each word as a list or vector of weights. Depending on the algorithm, the degree to which the individual weights can be interpreted varies substantially. Also, the algorithms differ in terms of how much the quality of the resulting vectors depends on the size of the provided collection of documents. 
Table~\ref{tbl:methodcomparison} is aimed at helping the practitioner to find the right method for the task at hand.\footnote{It should be emphasized that this comparison is only indicative. For details on comparison see, for example, \citet{edgar2016comparative,cimiano9explicit}.} All of the methods covered in Table~\ref{tbl:methodcomparison} are briefly described in the following text.

\begin{table}[htbp]
\caption{Methods generating word vectors}
\centering
\begin{tabular}{llll}
\toprule
Method & Required data size  & Features & Algorithmic approach \\ 
\midrule
BoW & small & Explicit (terms) & Term-document matrix \\
ESA & medium  & Explicit (documents) & Inverted index \\ 
LDA & smaller & Latent topics & Generative model \\ 
LSA & smaller & Latent concepts & Matrix factorization \\ 
word2vec & large  & Uninterpretable & Neural network \\ 
Glove & large & Uninterpretable & Regression model \\ 
\bottomrule
\end{tabular}
\label{tbl:methodcomparison}
\end{table}

\paragraph{Bag of Words (BoW)}
One of the most commonly applied type of vector space model is based on a \emph{term-document matrix}, where  rows correspond to terms (typically words) and columns to  documents. For each term, the matrix expresses the number of times it appears in the given document. This representation is also called a \emph{bag of words}. The term frequencies (TFs) act as weights that represent the degree to which the given word describes the document.  To improve results, these weights are further adjusted through normalization or through computing inverse document frequencies (IDFs) in the complete collection. IDF reflects the observation that  rarer terms -- those that appear only in a small number of documents -- are more useful in discriminating documents in the collection from each other than terms that tend to appear in all or most documents.
Bag-of-words representation incorporating IDF scores is commonly referred to as TF-IDF.

\paragraph{Semantic Analysis}
The \emph{explicit semantic analysis (ESA)} approach \citep{Gabrilovich07computingsemantic}  represents a particular word using a weighted list of documents (typically Wikipedia articles). ESA represents words based on an \emph{inverted index}, which it builds  from documents in the provided knowledge base.\footnote{ESA assumes that documents in the collection form a knowledge base -- each document covers a different topic.}
Each dimension in a word vector generated by ESA corresponds to a document in the training corpus, and the specific weight indicates to what extent that document represents the given word.

\emph{Latent semantic analysis} (LSA) \citep{landauer1997solution} and \emph{latent Dirichlet allocation} (LDA) \citep{blei2003latent}  are two  older, well-established algorithms, which are often used for topic modeling, namely, the identification of topics or concepts best describing a given document in the collection. The concepts and topics produced by these methods are \emph{latent}. That is, LDA topics are  not given an explicit label by the method (such as ``finances''), but instead can be interpreted through weights of associated words (such as ``money'' or ``dollars'' \citep{chen2016practical}).

\paragraph{Semantic Embeddings}
\emph{Word2vec} \citep{mikolov2013distributed} is a state-of-the-art approach to generating word vectors.  
The previously covered algorithms  generate interpretable word vectors essentially based on analyzing counts of occurrences of words. A more recent approach is based on predictive models. These use a predictive algorithm -- word2vec uses a neural network -- to predict a word given a particular context or vice versa. Word vectors created by word2vec (and related algorithms) are sometimes called \emph{word embeddings}: an individual word  is represented by a list of $d$ weights (real numbers).  

\emph{Glove} (Global Vectors for Word Representation) \citep{pennington2014glove} is an algorithm inspired by word2vec, which uses a weighted least squares model trained on
global word-word co-occurrence counts. Word embeddings trained by the Glove algorithm do particularly well on the word analogy tasks, where the goal is to answer questions such as ``Athens is to Greece
as Berlin is to $\_\_\_\_?$''



\begin{figure}
  \includegraphics[width=\linewidth]{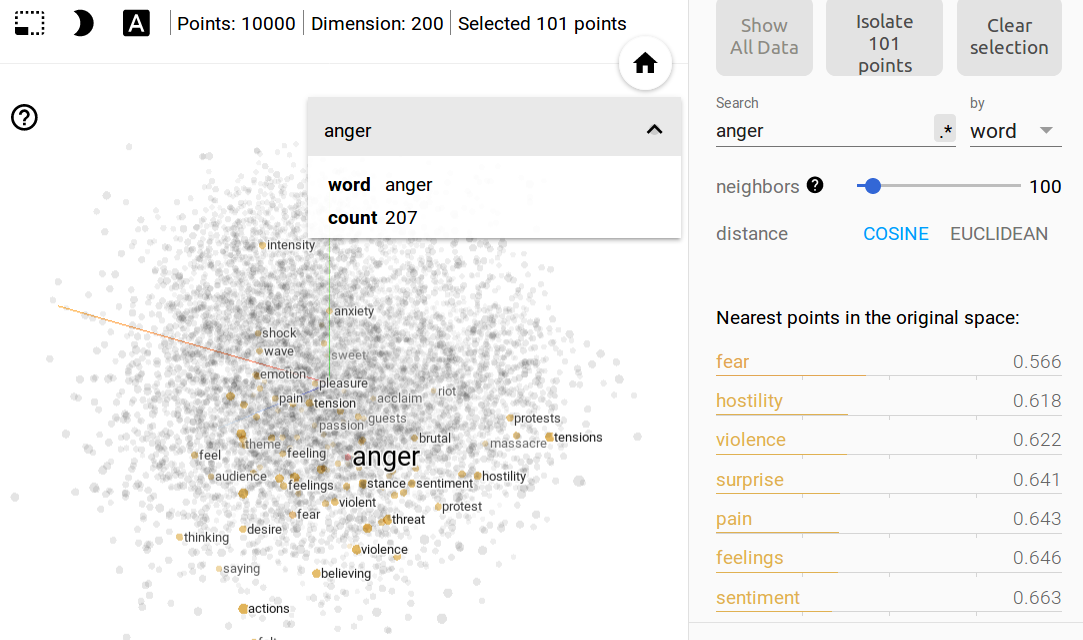}
  \caption{Nearest words to word "anger" (Embeddings Projector, Word2Vec 10K model)}
  \label{fig:projector}
\end{figure}

\paragraph{Quality of Results vs Interpretability of Word Vectors.}
Predictive algorithms such as word2vec have been shown to provide better results than models based on analyzing counts of co-occurrence of words across a  range of lexical semantic tasks, including word similarity computation \citep{baroni2014don}.  
While the individual dimensions in word2vec or Glove models do not directly correspond to explicit words or concepts as in ESA, distance between word vectors can be computed to find analogies and compute word similarities (see~Figure~\ref{fig:projector}).

\paragraph{\bf Applications in Behavioral Sciences}
\citet{caliskan2017semantics} have shown that semantic association of words measured using the distance of their embeddings generated by the Glove algorithm can reproduce results obtained with human subjects using the implicit association test. The results suggest that implicit associations might be partly influenced by similarities of words that co-occur with concepts measured by the implicit association test. The method could also be fruitful in predicting implicit associations and examining possible associations of people in the past.
Word embeddings might also be useful for the preparation of stimuli in tasks where semantic similarity of words is important, such as in semantic priming or memory research. The method provides a means of creating stimuli and also can be used to measure semantic similarity in models of performance on tasks depending on the semantic similarity of words. For example, \citep{howard2002does} used LSA to examine how semantically similar words recalled in sequence in a memory study. Similarly, the Deese-Roediger-McDermott paradigm \citep{roediger1995creating} uses semantically related words to elicit false memories. The described methods could then be used to measure the semantic similarity of words, which could influence the probability or strength of the false memories.

The LDA algorithm is typically used for topic modeling.
Based on an analysis of input documents, these algorithms  generate a list of topics. Each document is assigned a list of scores that expresses to what degree the  document corresponds to each of the topics.  Recent uses of LDA and word2vec include  a detection of fake news on Twitter \citep{8508520}.
For other examples of uses of the LSA and LDA algorithms  in a psychological context, we refer the reader to \citet{chen2016practical,edgar2016comparative}.







\subsection{Text Annotation} 
Textual documents can be extended with an additional structure using a variety of algorithms developed for natural language processing. 

\paragraph{Syntactic Parsing}
Analysis of a textual document often starts with syntactic tagging. This breaks the words in the input text into tokens and associates tokens with tags, such as parts of speech and punctuation. Syntactic parsing may also group tokens into larger structures, such as noun chunks or sentences.  Other types of tags include punctuation. Syntactic parsing may also group tokens into larger structures, such as noun chunks or sentences. Other types of processing include lemmatization --- reducing the different forms of a word to one single form --- which is important particularly for inflectional languages, such as Czech.

The result of syntactic parsing is typically used in further linguistic processing, but it also serves as a source of insights on the writing style of a particular group of subjects  \citep{o2017linguistic}.

\paragraph{Named Entity Recognition (NER)}
Syntactic parsing can already output noun chunks, such as names consisting of multi-word sequences (``New York'').
Named entity recognition goes one step further, by  associating each of these noun chunks with an \emph{entity type}. The commonly recognized types of entities are  \citep{tjong2003introduction}: persons, locations, organizations, and miscellaneous entities that do not belong to the previous three groups.


NER systems  are pretrained on large tagged textual corpora and are thus generally language dependent. Adjusting them to a different set of target classes requires a substantial amount of resources, particularly of tagged training data. 

\paragraph{Wikification: Linking Text to Knowledge Graphs}
The NER results are somewhat limited in terms of the small number of types recognized and lack of additional information on the entity. A process popularly known as  wikification addresses these limitations by linking entities to external knowledge bases. The reason why this process is sometimes called \emph{wikification} \citep{mihalcea2007wikify} is that  multiple commonly used knowledge bases are derived from Wikipedia.


The first step in entity linking is called mention detection. The algorithm identifies parts of the  input text, which can be linked to an entity in the domain of interest. For example, for input text ``Diego Maradona scored a goal'',  mention detection will output ``Diego Maradona'' or the corresponding positions in the input text. 

When mentions have been identified, the next step is their linking to the knowledge base. One of the computational challenges in this process is the existence of multiple matching entries in the knowledge base for a given mention. For example, the word ``Apple'' appearing in an analyzed Twitter message can be  \emph{disambiguated} in Wikipedia to Apple\_Inc or Apple (fruit). 

\begin{table}[htbp]
\centering
\begin{tabular}{llp{2cm}llrr}
\hline
URI & support & types & surfaceForm & offset & sim & perc \\ \hline
Apple\_Inc. & 14402 & Organisation, Company, Agent & Apple Inc. & 5 & 1.00 & 2.87E-06 \\ \hline
Steve\_Jobs & 1944 & Person, Agent & Steve Jobs & 27 & 1.00 & 8.66E-11 \\ \hline
ITunes & 13634 & Work, Software & iTunes & 53 & 0.98 & 2.12E-02 \\ \hline
\end{tabular}
\caption{Wikification example. ``Late Apple Inc. Co-Founder Steve Jobs 'Testifies' In iTunes Case'' generated by DBpedia Spotlight. The column names have the following meaning.  URI: values were stripped of the leading \url{http://dbpedia.org/resource/},  support: indicates how prominent is the entity by the number of inlinks in Wikipedia, types: were stripped of the leading \url{http://dbpedia.org/ontology/}, surfaceForm: the entity as it appears in the input tweet, offset: the starting position of the text in the input tweet in characters, sim: similarity between context vectors and the context surrounding the surface form, perc (percentageOfSecondRank): indicates confidence in disambiguation (the lower this score, the further the first ranked entity was "in the lead"). }
\label{tbl:spotlight}
\end{table}

Always assigning the most frequent meaning of the given word has been widely adopted as a base line in word sense disambiguation research \citep{navigli2009word}. When entity linking is performed, the knowledge base typically provides a machine-readable entity type, which might be more fine-grained than the type assigned by NER systems.  An example of a wikification output is shown in Table~\ref{tbl:spotlight}.

\paragraph{Entity Salience and Text Summarization}
When text is represented by entities, an optional processing step is to determine the level of salience  of the entity in the text. Entities with high salience can help to summarize content of longer documents, but the output of entity salience algorithms can also serve as input for subsequent processing, such as document classification.

Supervised entity salience algorithms, such as  the one described by \citet{Gamon2013}, are trained on a number of features derived from the entity mention (whether the word starts with an upper-case or lower-case letter), from the local context (how many characters the entity is from the beginning of the document), and global context (how frequently does the entity occur in inlinks and outlinks). Knowledge bases can be used as a complementary source of information \citep{dojchinovski2016crowdsourced}.

\paragraph{Sentiment Analysis}
With the proliferation of applications in social media, the analysis of sentiment and related psychological properties of text gained in importance. Sentiment analysis encompasses multiple tasks, such as determining valence and intensity of sentiment, determination of subjectivity, and detection of irony \citep{serrano2015sentiment}.

Most systems rely on lexicon-based analysis, machine learning, or  a combination of both approaches. Lexicon-based approaches rely on the availability of lists of words, terms, or complete documents, which are preclassified into different categories of sentiment. A well-known example developed for psychometric purposes is the LIWC2015 Dictionary, which assigns 6,400 words into several dozen nuanced classes  such as swear words, netspeak, or religion \citep{pennebaker2015development}.

\paragraph{\bf Applications in Behavioral Sciences}
Entities linked to knowledge graphs can be used to improve the results of many natural language processing tasks. \citet{troisi2018big}, for example, studied variables influencing the choice of a university by using wikification to find topics discussed in the context of writing about universities in various online sources. External information can be particularly useful in domains where the available documents are short and do not thus contain much information. To this end, \citet{varga2014linked} report significant improvement in performance when the content of tweets is linked to knowledge graphs as opposed to lexical-only content contained in  the input tweets.

The LIWC system has been widely used in the behavioral sciences (see the article  \citep{donohue2014validating}). Among other topics, it has been used to study close relationships, group processes, deception, and thinking styles \citep{tausczik2010psychological}. In general, it can be easily used to study differences in the communication of various groups. For example, it was used to analyze psychological differences between Democrats and Republicans by \citet{sylwester2015twitter}. This research  focused on general linguistic features, such as part of speech tags and sentiment analysis. 
The study found, for example, that those who identified as Democrats more commonly used first-person singular pronouns, and that the expression of positive emotions was positively correlated with following Democrats, but not Republicans.

Many uses of sentiment analysis deal with microposts such as Twitter messages.  Examples of this research include characterization of debate performance \citep{Diakopoulos:2010:CDP:1753326.1753504} or analysis of polarity of posts \citep{Speriosu:2011:TPC:2140458.2140465}.

\subsection{Document classification}
Document classification is a common task performed on top of a vector space representation of text, such as bag of words, but document classification algorithms can also take advantage of entity-annotated text \citep{varga2014linked}.
The goal of document classification is to assign documents in a given corpus to one of the  document categories. The training data consist of documents for which the target class is already known and specified in the input data.

In the following, we describe a centroid-based classifier, a well-performing algorithm. Next, we cover a few additional algorithms  and tasks.

\emph{Centroid Classifier}  \citep{han2000centroid}.
The centroid classifier  is one of the simplest classifiers working on top of the BOW representation. Input for the training phase is a set of documents for each target class, and the output is a \emph{centroid} for each category.
Centroid is a word vector, which is intended to represent the documents in the category. It is computed as an average of word vectors of documents belonging to the category. 

The application of the model works as follows. For each test document with an unknown class, its similarity to all target classes  is computed using a selected similarity measure. The class with the highest similarity is selected.
There are several design choices  when implementing this algorithm, such as the word weighting method, document length normalization, and the similarity measure.  The  common approach to the first two choices  is TF-IDF, covered in the ``Word Vectors and Word Embeddings'' subsection, and L1 normalization. L1 normalization is performed by dividing each element  in the given vector by the sum of absolute values of all elements in the vector. The similarity measure used for document classification is typically cosine similarity.


\paragraph{Other Tasks and Approaches}
The centroid classifier is a simple approach, which has the advantage of good interpretability. The simplicity of the algorithm can make it a good choice for large datasets. Centroid-based classifiers are noted to have excellent performance on multiple different  collections of documents but are not suitable for representing  classes that contain fine-grained subclasses \citep{pang2015dmkd}.



Support vector machines (SVM) \citep{boser1992training} is a frequently used algorithm for text classification, which can be adapted for some types of problems where centroid-based classification cannot be reasonably used. According to experiments reported by \citet{pang2015dmkd}, SVM is  a recommended algorithm for large \emph{balanced corpora}.  Balanced corpora have a similar proportion of documents belonging to individual classes.  SVMs can also be adapted to \emph{hierarchical classification}, where target classes can be further subdivided into subclasses \citep{dumais2000hierarchical}. 
Another adaptation of the text classification problem is \emph{multilabel text classification}, where a document is assigned multiple categories. 

\paragraph{\bf Applications in Behavioral Sciences}
Document classification methods have varied uses. One possible use is in predicting a feature of a person based on a text they wrote. For example, using a training set of documents, it is possible to train a model  to distinguish between documents written by men and women. Given a document for which an author is not known, the algorithm may be able to say whether the document was more likely to be written by a man or a woman. Similarly, in \citep{FLAIRS124417}, SVM and Bayes classifiers were used to  identify persona types based on word choice. Profiling using SVMs was also successfully applied for distinguishing among fictional characters \citep{D15-1208}.

The use of document classification can be further extended. Once the model is trained to classify documents using a list of features, it is possible to study and interpret the distinguishing features themselves. That is, it might be of interest not only to be able to predict the gender of the author of a document but also to say what aspects of the  documents written by males and females differ.


\section{External Knowledge Sources}
\label{sec:kg}
Enrichment with external knowledge can  be used to improve results of  machine learning tasks, but the additional information can also help to gain  new insights  about the studied problem \citep{paulheim2018machine}.

Two major types of knowledge sources for the machine learning tasks covered in this article  are knowledge graphs and lexical databases. In this section, we cover DBpedia and Wikidata,  prime examples of knowledge graphs which are semi-automatically generated from Wikipedia. For lexical databases, we cover WordNet, expert-created thesaurus with thousands of applications across many disciplines.

\subsection{Knowledge graphs}
Resources providing a mix of information in a structured and unstructured format are called \emph{knowledge bases}. A knowledge base can be called a \emph{knowledge graph} when information contained in it has a network structure and can  be obtained with structured queries.\footnote{
\url{https://www.ontotext.com/knowledgehub/fundamentals/what-is-a-knowledge-graph/}}
There is no universal \emph{graph query language} used to obtain information from knowledge graphs, but the openly available knowledge graphs covered in this section support SPARQL \citep{harris2013sparql}.  
The goal of a typical query is to retrieve a list of entities along with their selected properties, given a set of conditions. \emph{Entity}  roughly corresponds to a thing in human knowledge described by the knowledge graph. 


\emph{DBpedia}\footnote{\url{https://dbpedia.org}} \citep{lehmann2015dbpedia} is one of the largest and oldest openly available knowledge graphs.
The English version of DBpedia covers more than 6 million entities, but it is also available for multiple other languages.
For a knowledge base to contain the information on an entity, it must have been previously \emph{populated}. DBpedia is populated mostly by algorithms analyzing semistructured documents (Wikipedia articles).

\emph{Wikidata}\footnote{\url{https://wikidata.org}} \citep{vrandevcic2014wikidata} is another widely used knowledge graph, which is available since 2012. Wikidata currently contains information on 45 million items or entities.
Similar to DBpedia, Wikidata is partly populated by robots extracting data from Wikipedia, but it also allows the general public to contribute.

Information from DBpedia and Wikidata can be obtained either through a web interface, with a SPARQL query, or by downloading the entire knowledge graph.
 
\paragraph{Other Knowledge Graphs}
Thanks to the use of global identifiers for entities and their properties, many knowledge graphs are connected to the Linked Open Data Cloud. A list of more than 1,000 knowledge graphs cataloged by domain -- such as life sciences, linguistics, or media -- is maintained at \url{https://lod-cloud.net/}. 

In addition to open initiatives, there are proprietary knowledge graphs, which can be accessed via various APIs. These include Google Knowledge Graph Search API,  Microsoft's Bing Entity Search API, and Watson Discovery Knowledge Graph.

\paragraph{\bf Applications in Behavioral Sciences}
One of the main uses of Knowledge graphs in the behavioral sciences is in the study of the spread of disinformation \citep{disPlosOne,Fernandez:2018:OMC:3184558.3188730}. DBpedia is used for computational fact-checking in several systems, including DeFacto \citep{GERBER201585}.
Knowledge graphs are also used to enhance understanding of the text by linking keywords and entities appearing in text to more general concepts. DBpedia has  also been used to analyze the discourse of extremism-related content, including a detection of offensive posts \citep{vu34613,DBLP:conf/mmm/CompanyW19,radicalisationdetection}.


\subsection{WordNet and Related Lexical Resources} 
WordNet is a large English thesaurus that was created at  Princeton University \citep{fellbaum2010wordnet}. It covers nouns, verbs, adjectives, and adverbs. Synonyms are grouped together into \emph{synsets}, that is, sets of synonyms. In WordNet 3.0, there are about 150,000  words grouped into more than 100,000  synsets.
For each synset, there is a short dictionary explanation available called a \emph{gloss}. There are several types of relations captured between synsets depending on the type  of  synset, such as  hypo-hypernymy, antonymy, or holonymy-meronymy.  For example, for the noun ``happiness'' wordnet returns the synonym ``felicity'' and for ``sad'' the antonym ``glad''.

\paragraph{Use for Word Similarity Computation}
WordNet is also an acclaimed lexical resource that is widely used in the literature for word similarity and word disambiguation computations. 
With Wordnet,  one  can algorithmically   compute semantic similarity between a word and one or more other words. There are many algorithms -- or formulas -- for this purpose, which differ predominantly in the way they use the paths between the two words in the WordNet thesaurus as well as in the way they use external information -- such as how rare the given word is in some large collection of documents. Well-established algorithms include Resnik \citep{resnik95using_short}  and Lin \citep{Lin:1998:IDS:645527.657297}  measures.  A notable example in the behavioral context is the Pirro and Seco measure \citep{Pirro:2008:DIE:1483848.1483883}, which is inspired  by the feature-based theory of similarity proposed by \citet{Tversky77}. 

\paragraph{Use for Sentiment Analysis}
Further elaborating on the variety of possible uses of WordNet, recent research has provided an extension called ``Wordnet-feelings'' \citep{siddharthan2018wordnet}, which  assigns more than 3,000 WordNet synsets into nine categories of feeling. A related resource used for sentiment classification is SentiWordNet \citep{baccianella2010sentiwordnet}.

\paragraph{\bf Applications in Behavioral Sciences}

WordNet is often used in the behavioral sciences to complement free association norms, which are costly and time-consuming to develop \citep{maki2006efficient}.  \citet{maki2004semantic} showed that semantic distance computed from WordNet is related to participants' judgment of similarity.


Specific uses of WordNet in behavioral research include studies of perceptual inference \citep{johns2012perceptual}, access to memory \citep{buchanan2010access}, and predicting survey responses \citep{arnulf2014predicting}. For example, \citet{arnulf2014predicting} showed that semantic similarity of items computed with an algorithm using WordNet predicted observed reliabilities of scales as well as associations between different scales.

\section{Related Work}
\label{sec:rw}
In this section, we point readers to several works that also aimed at communicating recent advances in  machine learning algorithms and software to researchers in behavioral science. 
\citet{EDM-Behavioral} provide an edited volume exploring many topics and applications at the intersection of exploratory data mining and the behavioral sciences. Methodologically, the book has a strong focus on decision tree learning, exploring its use in areas as diverse as life-course analysis, the identification of academic risks, and clinical prediction, to name but a few.

\citet{tonidandel2018big} provide a discussion of ``big data'' methods applicable to  the organizational science, which is complemented by a list of various software systems across different programming languages (Python, R, ...), environments (cloud, desktop), and tasks  (visualization, parallel computing, ...). 
\citet{varian2014big}  reviews selected ``big data'' methods in the context of econometrics, focusing on random forests and trees. 


\citet{chen2016practical} give a practical introduction to "big data" research in psychology, providing an end-to-end guide covering topics such as a selection of a suitable database and options for data acquisition and preprocessing, focusing on web-based APIs and processing HTML data. Their article focuses on methods suitable for text analysis, giving  a detailed discussion, including worked examples for selected methods (LSA, LDA). There is also  a brief overview of the main subtasks in data mining, such as classification or clustering. The article also contains advice on processing large datasets, referring to the MapReduce framework.

\paragraph{Machine Learning vs.\ Big Data}
While many articles often use the term ``big data'', most data sets in behavioral science would not qualify.  According to  \citet{kitchin2017big,gandomi2015beyond}, big data consist of terabytes or more of data. Consequently, ``big data'' requires adaptation of existing algorithms so that they can be executed in a parallel fashion in a cloud or in grid-based computational environments. R users have the option to use some of the R packages for high-performance computing.\footnote{\url{https://cran.r-project.org/web/views/HighPerformanceComputing.html}}
Examples of dedicated big data architectures include Apache Spark or  cloud-based machine learning services \citep{hashem2015rise}.

\paragraph{Machine Learning as a Service}
In this article, we focused on packages available in the R ecosystem.\footnote{For a general introductory reference to R, we refer, e.g., to \citet{DataMiningR}.} The R data frame, used usually to store research data, is principally limited to processing data that do not exceed the size of available memory \cite[p.~399]{lantz2015machine}, which puts constraints on the size of analyzed data for packages that use this structure. As noted above, there are several options for scaling to larger data, but the behavioral scientist may find it most convenient to use a cloud-based machine learning system, such as BigML.\footnote{\url{https://bigml.com}}

MLaaS systems provide a comfortable web-based user interface, do not require installation or programming skills, and can process very large datasets. 
The disadvantage of using API-based or web tools such as MLaaS include impeded reproducibility of studies which used them for analysis. The researcher reproducing the analysis may not be able to employ the specific release of the system that was used to generate the results. The reason is that these systems are often updated.

\section{Conclusion}
The continuing shift of communication and interaction channels to online media provides a new set of challenges and opportunities for behavioral scientists. The fact that much interaction is performed online also allows for evolution in research methods.  For example, certain research problems may no longer require costly laboratory studies as suitable data can be obtained from logs of interactions automatically created by social networking applications and web sites. This article aimed to introduce a set of methods that allow for analyses of such data in a transparent and reproducible way. Where available, we therefore suggested software available under an open source license. 

We put emphasis on selecting proven algorithms, favoring those that generate interpretable models that can be easily understood by a wide range of users. When easy-to-interpret models lead to worse results than more complex models, it is possible to use the latter to improve the former. For example, \citet{agrawal2019using} used neural networks to predict moral judgments. Because the neural network model was itself not easily interpretable, they looked at situations where the neural network model fared particularly well in comparison to a simpler, but more easily interpretable, choice model. They then iteratively updated the choice model to better predict judgments in situations where the neural network model predicted better. A similar strategy can be used generally by behavioral scientists if the interpretability of the models is considered valuable.

There are several other noteworthy areas of machine learning that could be highly relevant to particular subdomains of behavioral science. We left them uncovered due to space constraints. These include reinforcement learning, image processing, and the discovery of interesting patterns in data. Another interesting technological trend in terms of how data are collected and processed is the connection between crowdsourcing services and Machine Learning as a Service offering. Crowdsourcing may decrease the costs by outsourcing some parts of research such as finding and recruiting participants and can also aid replicability by engaging large and varied participant samples. 

Employment of MLaaS systems may have benefits in terms of setup costs, ease of processing,  and the security of the stored data. On the other hand, experimenters relying on crowdsourcing lose control of the laboratory environment. MLaaS may impede reproducibility and accountability of the analysis since the results of these systems may vary in time as they are often updated. See article by \citet{crump2019conducting}, which is in this issue, on the challenges of recruiting participants.  

Overall, we expect that the largest challenge for the behavioral scientist in the future will not be the choice or availability of suitable machine learning methods. More likely, it will be ensuring compliance with external constraints and requirements concerning ethical, legal, and reproducible aspects of the research.

\section*{Acknowledgments}
The authors disclosed receipt of the following financial support for the research, authorship, and/or publication of this article: TK was supported by Faculty of Informatics and Statistics, University of Economics, Prague by grant IGA 33/2018 and by institutional support for research projects. TK would like to thank BigML Inc. for providing a subscription that allowed to test processing of large datasets in BigML.com free of charge. The work of \v{S}B was supported by the Internal Grant Agency of the Faculty of Business Administration, University of Economics, Prague (Grant No. IP300040).

\bibliography{references,tomas,decision_trees,rules,kdd,jf,ml,ensembles,preferences,plbook,web,nn}

\end{document}